\documentclass[journal]{vgtc}                     

\onlineid{1599}

\vgtccategory{Research}

\vgtcpapertype{please specify}

\newcommand{\sysname}{SurroFlow} 

\title{{\sysname}: A Flow-Based Surrogate Model for Parameter Space Exploration and Uncertainty Quantification}

\author{%
  \authororcid{Jingyi Shen}{0000-0001-5478-3993},
  Yuhan Duan, and 
  Han-Wei Shen
}

\authorfooter{
  \item Jingyi Shen, Yuhan Duan, and Han-Wei Shen are with The Ohio State University. E-mail: \{shen.1250\,$|$\,duan.418\,$|$\,shen.94\}@osu.edu\,
}

\abstract{%
Existing deep learning-based surrogate models facilitate efficient data generation, but fall short in uncertainty quantification, efficient parameter space exploration, and reverse prediction. In our work, we introduce {\sysname}, a novel normalizing flow-based surrogate model, to learn the invertible transformation between simulation parameters and simulation outputs. 
The model not only allows accurate predictions of simulation outcomes for a given simulation parameter but also supports uncertainty quantification in the data generation process. Additionally, it enables efficient simulation parameter recommendation and exploration. 
We integrate {\sysname} and a genetic algorithm as the backend of a visual interface to support effective user-guided ensemble simulation exploration and visualization. 
Our framework significantly reduces the computational costs while enhancing the reliability and exploration capabilities of scientific surrogate models. 
}


\graphicspath{{figs/}{figures/}{pictures/}{images/}{./}} 

\usepackage{tabu}                      
\usepackage{booktabs}                  
\usepackage{lipsum}                    
\usepackage{mwe}                       

\usepackage{mathptmx}                  

\usepackage{amsfonts}
\usepackage{multirow}
\usepackage{makecell}
\usepackage{amsmath}
\usepackage{fontawesome}
\usepackage{algorithm}
\usepackage[noend]{algpseudocode}

\begin{document}



\maketitle
\section{Introduction}
In fields like fluid dynamics, climate, and weather research, ensemble simulations have become critical for estimating uncertainty, improving decision-making, and enhancing the scientific understanding of complex systems. To fully study scientific phenomena and analyze the sensitivity of simulation parameters, scientists often need to conduct a sequence of simulations and compare the results generated from different configurations. 
However, scientific simulations can be computationally expensive and time-consuming. To speed up the simulation, analysis, and knowledge discovery cycle, a major research effort has been put into developing simulation and visualization surrogates to approximate the results of complex and computationally expensive simulations with much reduced costs. 

Currently, many deep learning-based surrogate models have been proposed in scientific visualization fields to assist parameter space exploration for ensemble simulations~\cite{he2019insitunet,shi2022vdl,gramacy2004parameter}. Although they have begun to gain traction, there still exist several limitations. 
The first limitation is the lack of uncertainty quantification for surrogate models. Surrogate models are approximations of the actual simulations and inherently contain uncertainties, it is important to convey uncertainties associated with the surrogate model's outputs so that scientists know how much they can trust the model's predictions. Uncertainty quantification also provides insights into how sensitive the surrogate models' predictions are with respect to the changes in the simulation parameters. This sensitivity information is crucial for scientists as it guides the exploration direction about where to focus more, for example, regions with high data variance, by running additional simulations. 
The second challenge is the lack of efficient exploration across the parameter space. Current works only support limited functionality such as predicting the data for a given simulation parameter set, but do not assist in discovering potential optimal simulation configurations across the large parameter space. Parameter space exploration based on heuristic or brute-force approaches can be computationally intensive and inefficient. A systematic and efficient way for parameter space exploration that considers scientists' interests is needed.
The third challenge is reverse prediction, particularly the ability to predict the simulation parameters that can produce a specific simulation result. This ability is crucial when scientists explore the optimal outcomes by directly manipulating the simulation outputs but lack the knowledge of the underlying simulation parameters that could generate those manipulated data. Thus, there is a pressing need for effective approaches that can infer simulation parameters from observed outcomes.

In this work, we propose {\sysname}, a surrogate model based on the normalizing flow~\cite{kingma2018glow,dinh2016density,rezende2015variational}. Normalizing flow is a type of generative model for modeling complex probability distributions. By learning the invertible transformation between a simple base distribution (e.g., Gaussian distribution) and a complex target distribution, we address the above challenges as follows. 
First, {\sysname} is a conditional normalizing flow model combined with an autoencoder, trained on pairs of simulation parameters and corresponding simulation outcomes. As it models the distribution of simulation data conditioned on the simulation parameters, {\sysname} can quantify the uncertainty in the surrogate modeling process. {\sysname} takes simulation parameters as the conditional input, by sampling from the simple base distribution multiple times and reconstructing them in the complex conditional distribution, the variations among the reconstructed outputs can reflect the uncertainties in the surrogate prediction. 
Second, the proposed uncertainty-aware surrogate model {\sysname} can assist interactive exploration of simulation parameters guided by scientists' interests and goals. This is done by integrating the trained {\sysname} with a genetic algorithm, empowered by an interactive visual interface for user-guided automatic parameter space exploration. Scientists can specify their preferences and optimization objectives via the visual interface. The genetic algorithm then iteratively explores generations of simulation parameters based on scientists' inputs, while {\sysname} operates in the backend to efficiently predict data and uncertainties for the given simulation parameters. Once scientists set up the objectives and trade-offs among data similarity, diversity, and uncertainty interactively on the visual interface, simulation parameter space exploration can proceed automatically. 
Third, another essential benefit of flow-based models is that they can perform prediction in both forward and reverse directions. 
In the forward direction, {\sysname} can predict data conditioned on the input simulation parameters, allowing for the efficient generation of high-quality synthetic data. Conversely, in the backward direction, {\sysname} can predict the underlying simulation parameters that produce the given simulation data. This bidirectional prediction allows an effective framework for both surrogate modeling and reverse prediction, enhancing the flexibility of parameter space exploration and post-hoc analysis. 

In contrast to other surrogate models in the visualization field ~\cite{he2019insitunet,shi2022gnn,shi2022vdl}, {\sysname} has made significant advancements by combining the conditional normalizing flows with autoencoders for efficient latent space modeling, allowing uncertainty quantification of the surrogate model’s outputs. {\sysname} also features an improved architecture for bi-directional predictions. Additionally, it integrates a genetic algorithm and an interactive visual interface for efficient simulation parameter space exploration.
Through qualitative and quantitative evaluations, we demonstrate the effectiveness of the proposed approach for automatic user-guided simulation parameter recommendation and exploration. 
In summary, the contributions of our work are:  
\begin{itemize}
\item First, we propose an invertible flow-based uncertainty-aware surrogate model. 
\item Second, we utilize a genetic algorithm that considers the similarity, diversity, and uncertainty of data during the automatic parameter space exploration process. 
\item Third, we employ an interactive visual interface for uncertainty analysis and provide scientists with effective control over simulation parameter space exploration. 
\end{itemize}
\section{Related Works}
We employ an invertible normalizing flow for surrogate modeling with uncertainty estimation. In this section, we provide an overview of related research on visualization surrogate models for scientific data, global sensitivity analysis techniques, and uncertainty estimation for deep neural networks.  

\textit{Surrogate Models for Parameter Space Exploration.}
In the fields of oceanography and climate science, scientists frequently run computational simulations~\cite{khater2023computational, meier2022oceanographic} to explore parameter spaces. 
Replacing these costly and time-consuming simulation processes, numerous surrogate models~\cite{razavi2012review,alizadeh2020managing,cozad2014learning} have been proposed. 
Among these, Gramacy et al.~\cite{gramacy2004parameter} construct non-stationary Gaussian process trees that adaptively sample within the input space, enabling the selection of more efficient designs. He et al.~\cite{he2019insitunet} introduce InSituNet, an image-based surrogate model designed for real-time parameter space exploration. 
However, its reliance on regular grid mappings limits its application to unstructured data. To overcome this limitation, Shi et al.~\cite{shi2022gnn} propose GNN-Surrogate, specifically designed for navigating simulation outputs on irregular grids. Shi et al.~\cite{shi2022vdl} also introduce the VDL-Surrogate model based on view-dependent neural-network latent representations. It employs ray casting from varied viewpoints to aggregate samples into compact latent representations, thereby optimizing computational resource use and enhancing high-resolution explorations. Similarly, Danhaive et al.~\cite{danhaive2021design} propose a surrogate model for performance-driven design exploration within parametric spaces using conditional variational autoencoders. They employ a sampling algorithm to distill a dataset that captures valuable design insights and employ the variational autoencoders as surrogates for the simulation process. 
However, existing learning-based works do not account for uncertainties within the surrogate model. To address this deficiency, we propose a flow-based surrogate to enable uncertainty quantification of the surrogate model's prediction process.

\textit{Global Sensitivity Analysis.} 
Global sensitivity analysis methods include variance-based, differential-based, and regression-based approaches. They are crucial for understanding how parameters influence model outcomes. 
Variance-based methods, also known as ANalysis Of VAriance (ANOVA), decompose the output's variance into a sum of contributions from inputs and their interactions. Based on the application, Sobol indices~\cite{sobol1990sensitivity} or Design of Experiment~\cite{LAMBONI2011DoE} are used to evaluate inputs' effects on the output.  
Differential-based methods, like the Morris~\cite{morris1991factorial}, calculate input effects by varying each factor individually while keeping others fixed. This method is computationally efficient and easy to implement.
Regression-based methods utilize linear approximations of the function to compute sensitivity based on metrics like Correlation Coefficients and Standardized Regression coefficients~\cite{Iooss2011SRC}. 
There are traditional surrogate models with global sensitivity analysis, such as Ballester et al.~\cite{ballester2019sobol}'s Sobol tensor train decomposition. While efficient, these surrogate methods are outperformed by deep neural networks.

\textit{Uncertainty in Deep Neural Networks.}
Deep neural networks are often viewed as black boxes due to unclear decision-making rules. For scientific visualization and analysis, uncertainty quantification helps measure the reliability of the model, offering scientists a level of confidence about the model's outputs. 
A large number of uncertainty quantification methods have been proposed, including probabilistic models (e.g., Bayesian neural networks~\cite{neal2012bayesian}, Deep Gaussian process~\cite{damianou2013deep}), ensemble methods (e.g., DeepEnsemble~\cite{lakshminarayanan2017simple}), and deep generative model-based methods~\cite{kingma2013auto,goodfellow2014generative,ho2020denoising}. In our work, we employ a deep generative model called normalizing flow~\cite{kingma2018glow,rezende2015variational,papamakarios2021normalizing,shen2023psrflow} as the surrogate model to quantify uncertainties.


\section{Background: Normalizing Flow}\label{sect:BG_NF}
{\sysname} is based on the normalizing flow for the conditional generation of simulation data conditioned on the simulation parameters.  
Normalizing flow models are a type of generative models that aim to learn a mapping from a simple probability distribution to a more complex one, allowing for the generation of high-quality data samples in the target distribution. 

\vspace{-5pt}
\begin{figure}[htp]
    \centering
    \includegraphics[width=\columnwidth]{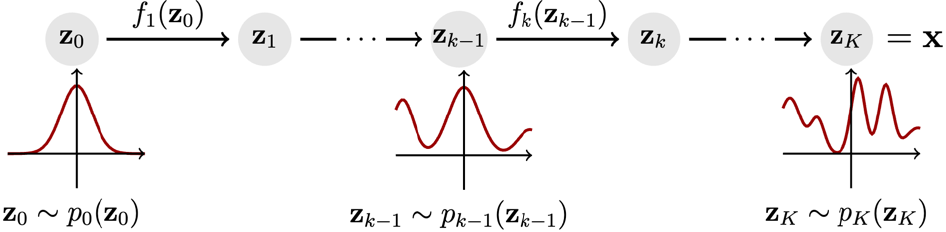}
    \vspace{-12pt}
    \caption{A normalizing flow turns a simple base distribution into a more complex one. This mapping is invertible in the opposite direction.}
    \label{fig:bg_nf}
    \vspace{-2pt}
\end{figure}

The key idea behind normalizing flow is to learn a series of invertible transformation functions that can gradually map samples from a simple distribution into a complex target distribution of interest, as shown in~\cref{fig:bg_nf}. Since the mapping is invertible, one can perform the transformation in the opposite direction, i.e., encode samples in the complex distribution into latent vectors with a simple and tractable distribution.
Formally, denote the observed variables as $\mathbf{x}$ and latent variables as $\mathbf{z}$, the mapping in the normalizing flow model, given by $f_\theta: \mathbb{R}^n \rightarrow \mathbb{R}^n$, is invertible such that $\mathbf{x}=f_\theta(\mathbf{z})$ and $\mathbf{z}=f_\theta^{-1}(\mathbf{x})$. 
The invertible mapping function $f_\theta$ consists of a sequence of bijective functions $f_\theta = f_{K} \circ f_{K-1} \circ \cdot \cdot \cdot \circ f_{1}$, so that
\begin{equation} \label{eq:bg_nf}
\mathbf{x} = \mathbf{z}_K = f_{K} \circ f_{K-1} \circ \cdot \cdot \cdot \circ f_{1}(\mathbf{z}_0),  
\end{equation}
\noindent where each $f_i$ (for $i=1, ..., K$) is learnable and invertible. With the stacked $K$ invertible functions, a normalizing flow can transform a latent variable $\mathbf{z}_0$, which follows the simple distribution (e.g., Gaussian distribution), into the target variable $\mathbf{z}_K$ with a more complex target distribution $p(\mathbf{x})=p_K(\mathbf{z}_K)$. 
Using the change of variables, the log-likelihood of $\mathbf{z}_K$ can be computed as
\vspace{-3pt}
\begin{equation} \label{eq:bg_logx}
\log p_K(\mathbf{z}_K) = \log p_0(\mathbf{z}_0) - \sum_{i=1}^{K} \log \left| \det{\frac{\partial f_i(\mathbf{z}_{i-1})} {\partial \mathbf{z}}_{i-1}} \right|.
\vspace{-2pt}
\end{equation}
Given the density of $\mathbf{z}_0$ and the learnable transformation functions $f_\theta$, one can compute the log-likelihood of any $\mathbf{x}$ through~\cref{eq:bg_logx}. 
The optimization goal for normalizing flows is to find the $\theta$ that maximizes the log-likelihood of training samples, computed by~\cref{eq:bg_logx}. 
The invertibility of normalizing flows is theoretically constrained by design. 
Various learnable and invertible functions for $f_i (i=1, ..., K)$ have been proposed, but the most efficient and simplest approach is affine coupling layers~\cite{dinh2016density}.
The input variable $\mathbf{z}$ of this coupling layer is split into two parts $\mathbf{z}_{1:j}$ and $\mathbf{z}_{j+1:d}$ at index $j$, where the first part is unchanged and is used to parameterize the affine transformation for the second part. The output variable $\mathbf{z}^\prime$ can be formulated as
\begin{equation} \label{eq:bg_affine}
\vspace{-2pt}
\begin{gathered} 
\mathbf{z}^\prime_{1:j} = \mathbf{z}_{1:j}, \\
\mathbf{z}^\prime_{j+1:d} = T(\mathbf{z}_{1:j}) + S(\mathbf{z}_{1:j}) \odot \mathbf{z}_{j+1:d}.
\end{gathered}
\end{equation}
\noindent The function $S(\cdot)$ and $T(\cdot)$ are neural networks and are not invertible. The sum and multiplication are element-wise operations. The complexity of $S(\cdot)$ and $T(\cdot)$ is essential for the modeling capability of coupling layers. Inverting the coupling layer can be performed by element-wise subtraction and division: 
\vspace{-2pt}
\begin{equation} \label{eq:bg_affine_inv}
\begin{gathered} 
\mathbf{z}_{1:j} = \mathbf{z}^\prime_{1:j}, \\
\mathbf{z}_{j+1:d} = (\mathbf{z}^\prime_{j+1:d} - T(\mathbf{z}_{1:j})) / S(\mathbf{z}_{1:j}).
\end{gathered}
\end{equation}

By using multiple invertible masking~\cite{dinh2016density, kingma2018glow}, normalization~\cite{kingma2018glow}, and coupling layers, we can ensure the input variables are fully processed and the model has enough capacity to learn the invertible transformations, enabling the model to transform between complex and simple distributions.

Normalizing flow models inherently provide a measure of uncertainty, as they can capture the complex posterior distribution of data. 
In our work, we employ normalizing flow models to capture the conditional distribution of simulation data conditioned on the simulation parameters, facilitating uncertainty quantification and efficient parameter space exploration for ensemble data. 


\section{Method}
Existing surrogate models lack uncertainty estimation in the prediction process and they only support predicting simulation results based on the simulation parameters specified by scientists. 
However, the critical issue often faced by scientists is the lack of knowledge about which simulation parameters are of interest. The exploration process can be computationally intensive and inefficient. This underscores the need for efficient user-guided exploration of the simulation parameter space. 
To address this gap, we propose an uncertainty-aware surrogate model {\sysname} and integrate it with a genetic algorithm to support efficient user-driven exploration of the simulation parameter space.

\subsection{Overview}

\vspace{-8pt}
\begin{figure}[htp]
    \centering
    \includegraphics[width=\columnwidth]{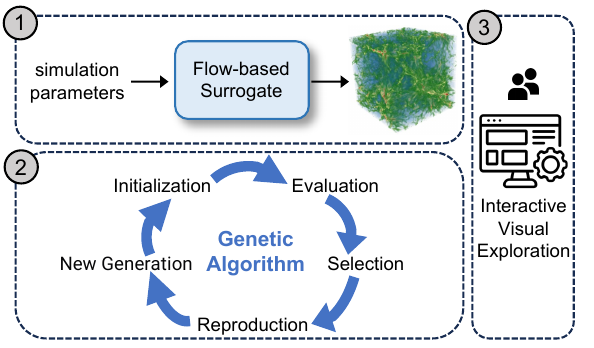}
    \vspace{-12pt}
    \caption{Overview of our approach. (1) An uncertainty-aware surrogate model is built to predict simulation outcomes for simulation parameters. (2) A genetic algorithm is utilized for efficient simulation parameter optimization. (3) Integrating the surrogate model and genetic algorithm into an interactive visual system for user-guided parameter space exploration.}
    \label{fig:overview}
    \vspace{-6pt}
\end{figure}

\Cref{fig:overview} shows an overview of our framework, which contains three major components. 
\textbf{First}, we propose {\sysname}, a normalizing flow-based surrogate model with uncertainty quantification ability. The model is trained on pairs of simulation parameters and simulation data. Once well-trained, scientists can utilize {\sysname} to generate data by taking simulation parameters as the conditional input. 
\textbf{Second}, to reduce the computational cost of intensive parameter space exploration for ensemble simulations, 
we cast exploration as a constrained optimization problem solved using a genetic algorithm. 
The constraints we consider are based on the preferences provided by scientists interactively.
\textbf{Third}, we integrate {\sysname} and the genetic algorithm as the backend for an interactive visual exploration system. With the visual interface, scientists can specify their interests and give indications about which exploration direction they are more interested in, allowing for efficient user-guided parameter space exploration considering simulation data similarity, diversity, and uncertainty.

\subsection{Uncertainty-Aware Surrogate Model}\label{sect:surrogate}
In this section, we discuss details about the proposed surrogate model {\sysname}, including (1) simulation data representation extraction, (2) conditional data generation and uncertainty quantification of the surrogate model, and (3) reverse prediction of simulation parameters for given simulation data. 

\subsubsection{Data Representation Extraction}\label{sect:ae}

\begin{figure}[htp]
    \centering
    \includegraphics[width=0.8\columnwidth]{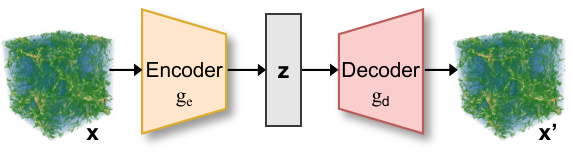}
    \vspace{-9pt}
    \caption{Autoencoder for dimensionality reduction. }
    \label{fig:ae}
    \vspace{-10pt}
\end{figure}

{\sysname} contains two parts that contribute to efficient data generation. The first part is an autoencoder model for 3D data reduction. 
Flow-based models require constant data dimensionality to maintain invertibility, which ensures that there exists a one-to-one correspondence between the input and output of each transformation. Despite normalizing flows' success in density modeling, training and testing these models on high-dimensional data is computationally expensive. 

To reduce the computational costs, we pre-train a 3D autoencoder to project the original high-dimensional data onto a lower-dimensional latent space. 
Given a simulation data $\mathbf{x} \in\mathbb{R}^{D \times H\times W}$, with dimensions $D$, $H$, and $W$ representing depth, height, and width, respectively, we train an autoencoder consisting of an encoder $g_e$ and a decoder $g_d$, both designed based on convolutional neural networks with residual connections~\cite{he2016deep}, to extract a compact latent representation $\mathbf{z}$. 
This encoding ensures the latent representation $\mathbf{z}$ captures the essential characteristics of the data $\mathbf{x}$ with reduced dimensionality, allowing the decoder to accurately reconstruct $\mathbf{x}^{\prime}$ from latent representation $\mathbf{z}$ with low loss compared to the original data $\mathbf{x}$. This process can be formulated as
\vspace{-1pt}
\begin{equation} \label{eq:ae}
\begin{gathered} 
\mathbf{z}=g_e(\mathbf{x}), \ \text{and} \
\mathbf{x}^{\prime}=g_d(\mathbf{z}).
\end{gathered} 
\vspace{-1pt}
\end{equation}

In the later stage, the normalizing flow model will be trained on the latent representations of data instead of their original high-dimensional counterparts. 
During inference, the flow-generated results will be further processed by the decoder of the autoencoder so that scientists can get the reconstructed data in the original data space. 

\subsubsection{Conditional Modeling for Uncertainty Quantification} \label{sect:surroflow}

In our work, we model the relationship between simulation parameters and simulation data as probabilistic distributions instead of deterministic mappings. 
This decision is motivated by two main factors. 
First, the neural network parameterized mapping between simulation parameters and data is an approximation, which inherently contains uncertainties. These uncertainties come from the limitations of the neural network in capturing the true complexity of the simulation relationships. 
Second, a probabilistic model is preferred since it is in general more robust. In most cases, since we do not have a large enough training dataset to approximate the complex but unknown mapping function, probabilistic modeling can better account for the range of possible outcomes. 

\begin{figure}[htp]
    \centering
    \includegraphics[width=0.9\columnwidth]{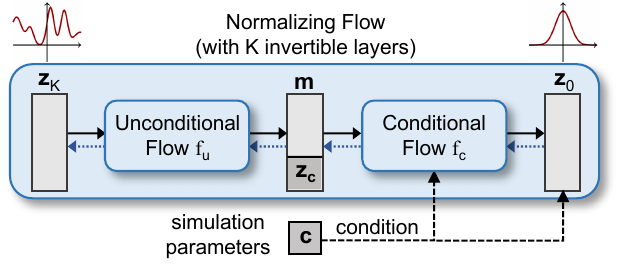}
    \vspace{-8pt}
    \caption{
    {\sysname} is a conditional normalizing flow for surrogate modeling. It is trained on pairs of simulation parameters ($\mathbf{c}$) and simulation data latent representations ($\mathbf{z}_K$). During inference, {\sysname} takes simulation parameters $\mathbf{c}$ as the input to sample $\mathbf{z}_0$ and outputs the reconstructed latent representation $\mathbf{z}_K$ (indicated by the blue dotted arrow). In the reverse direction, {\sysname} takes latent representations $\mathbf{z}_K$ as input and predicts the corresponding simulation parameters (black solid arrow).
    }
    \label{fig:nf}
    \vspace{-5pt}
\end{figure}

As shown in~\cref{fig:nf}, {\sysname} is a surrogate model based on a conditional normalizing flow to approximate the complex and time-consuming simulation process. {\sysname} learns a probabilistic mapping function of the simulation data conditioned on the simulation parameter. 
The learned probabilistic mapping will facilitate uncertainty quantification of the surrogate model. 
To reduce the computational cost, we train the normalizing flow in the latent space. This means the simulation data $\mathbf{x}$ used for training are first encoded into latent representations $\mathbf{z}$ by the trained encoder $g_e$, as described in~\cref{eq:ae}. 

Given pairs of simulation parameters and the latent representations of the corresponding simulation output, our goal is to model the conditional distribution $p(\mathbf{z} \mid \mathbf{c})$, where $\mathbf{c}\in\mathbb{R}^{n}$ denotes the condition (i.e., an n-dimensional vector representing a set of simulation parameters) and $\mathbf{z}$ denotes the latent representation of the corresponding simulation result. 
However, this conditional distribution $p(\mathbf{z} \mid \mathbf{c})$ is complex and intractable. To model this distribution, we introduce a latent variable $\mathbf{z}_0$ with a well-defined Gaussian distribution $p_0(\mathbf{z}_0 \mid \mathbf{c})$, and model the unknown conditional distribution $p(\mathbf{z} \mid \mathbf{c})$ by learning a transformation function $f$. 
This function $f$ transforms a sample $\mathbf{z}_0$ from the simple, known distribution $p_0(\mathbf{z}_0\mid \mathbf{c})$ to a sample $\mathbf{z}$ within the complex, target distribution $p(\mathbf{z} \mid \mathbf{c})$, formulated as
\vspace{-3pt}
\begin{equation} \label{eq:zk}
\mathbf{z} = \mathbf{z}_K = f(\mathbf{z}_0, \mathbf{c}),
\end{equation}
where $f$ is a normalizing flow, consisting of $K$ invertible transformation layers. $\mathbf{z}_K$, equal to the target complex data sample $\mathbf{z}$, is the resulting output after $K$ transformations of input $\mathbf{z}_0$.
This architecture is bidirectional so it can process data in both forward and reverse directions. For simplicity, we discuss how to transform data from a simple Gaussian distribution to a complex one. 
The latent representation $\mathbf{z}_0$ follows a Gaussian distribution conditioned on $\mathbf{c}$, described by
\vspace{-2pt}
\begin{equation} \label{eq:p0}
p_0(\mathbf{z}_0\mid \mathbf{c}) = \mathcal{N}(\mathbf{z}_0; \mu_c, \Sigma_c),
\end{equation}
\noindent where mean $\mu_c$ and variance $\Sigma_c$ of the Gaussian are predicted based on the condition $\mathbf{c}$ using two neural networks $h_{\mu}$ and $h_{\Sigma}$, i.e., $\mu_c = h_{\mu}(\mathbf{c})$ and $\Sigma_c = h_{\Sigma}(\mathbf{c})$.
Given the Gaussian distribution $p_0(\mathbf{z}_0\mid \mathbf{c})$ and the flow model $f$, based on the previous discussion in~\cref{sect:BG_NF} and~\cref{eq:bg_logx}, we can derive the log-likelihood of the target distribution $p(\mathbf{z}\mid \mathbf{c})$ as
\vspace{-2pt}
\begin{equation} \label{eq:pz}
\begin{gathered} 
\log p(\mathbf{z}\mid \mathbf{c}) = \log p_K(\mathbf{z}_K\mid \mathbf{c}) \\ 
= \log p_0(\mathbf{z}_0\mid \mathbf{c}) - \sum_{i=1}^{K} \log \left| \det{\frac{\partial f_i(\mathbf{z}_{i-1} \mid \mathbf{c})} {\partial \mathbf{z}_{i-1}}} \right|.
\vspace{-2pt}
\end{gathered} 
\end{equation}
This log-likelihood will be used to optimize the normalizing flow.

The $K$ invertible transformation layers in the flow $f$ are divided into two groups, i.e., $K_1$ unconditional layers and $K_2$ conditional layers ($K=K_1 + K_2$). 
As illustrated by the blue dashed arrows in~\cref{fig:nf}, the latent representation $\mathbf{z}_0$ will first go through a \textbf{conditional} normalizing flow $f_{c}$ with $K_1$ layers whose transformations depend on the conditional variable $\mathbf{c}$, enabling the modeling of the distribution externally conditioned on the simulation parameter. 
After the conditional part, the second component is the \textbf{unconditional} normalizing flow $f_{u}$ with $K_2$ invertible transformation layers. These layers take no conditional input and simply transform the input into the final complex target distribution. 
We formulate this process as
\vspace{-3pt}
\begin{equation} \label{eq:condflow}
\begin{gathered} 
\mathbf{m} = f_{c}(\mathbf{z}_0, \mathbf{c}), \\
\mathbf{z}_K = f_{u}(\mathbf{m}),
\end{gathered}
\vspace{-2pt}
\end{equation}
\noindent where $\mathbf{m}$ is the transformation result of $\mathbf{z}_0$ after going through the conditional flow $f_c$. The output of the unconditional flow $f_u$ is $\mathbf{z}_K$. Once well-trained, $\mathbf{z}_K$ should match the latent representation of the corresponding simulation data for a given parameter $\mathbf{c}$. In other words, {\sysname} can transform a base Gaussian distribution into a target one that represents the actual simulation output. 

The explicit conditional distribution modeling allows not only efficient sampling but also uncertainty quantification of the data generation process. 
Uncertainty quantification is crucial for scientific visualization and analysis of surrogate models since it will provide scientists with information about the quality of results. 
Given a conditional input, scientists can efficiently sample the Gaussian latent space for high-quality surrogate data generation. 
The Gaussian distribution's variance reflects the uncertainty of the surrogate model, stemming from its approximation error of the complex simulation process.
By sampling the Gaussian latent space and assessing variations in the reconstruction, scientists can quantify the uncertainties associated with the data produced by the surrogate model. We summarize this process in~\cref{alg:uq}.

\vspace{-2pt}
\begin{algorithm}
\caption{Surrogate prediction with uncertainty quantification}\label{alg:uq}
\begin{algorithmic}[1]
\State \textbf{Input:} Simulation parameter $\mathbf{c}$, number of samples $n$
\State \textbf{Output:} Predicted data $\bar{\mathbf{x}}$, quantified uncertainties $\mathbf{x}_{var}$
\Procedure{PredictAndQuantify}{$\mathbf{c}$, $n$}
    \State $X \gets$ initialize an empty list to save $n$ outputs
    \State $\mu_c \gets h_{\mu}(\mathbf{c})$, $\Sigma_c \gets h_{\Sigma}(\mathbf{c})$
    \State $p_0(\mathbf{z}_0\mid \mathbf{c}) \gets \mathcal{N}(\mathbf{z}_0; \mu_c, \Sigma_c)$
    \For{$i = 1$ to $n$}
        \State $\mathbf{z}_0 \gets$ sample from Gaussian distribution $p_0(\mathbf{z}_0 \mid \mathbf{c})$
        \State $\mathbf{m} \gets f_c(\mathbf{z}_0, \mathbf{c})$, apply conditional flow $f_c$ on $\mathbf{z}_0$
        \State $\mathbf{z} \gets f_u(\mathbf{m})$, apply unconditional flow $f_u$ on $\mathbf{m}$
        \State $\mathbf{x}^{\prime} \gets g_d(\mathbf{z})$, use decoder $g_d$ to get reconstructed data
        \State $X \gets X \cup \{\mathbf{x}^{\prime}\}$
    \EndFor
    \State $\bar{\mathbf{x}} \gets$ Mean($X$)
    \State $\mathbf{x}_{var} \gets$ Variance($X$)
    \State \textbf{return} $\bar{\mathbf{x}}$, $\mathbf{x}_{var}$
\EndProcedure
\end{algorithmic}
\end{algorithm}
\vspace{-8pt}

\subsubsection{Reverse Prediction} \label{sect:reverse_pred}
To enable the reverse prediction of simulation parameters from simulation data, we introduce a constraint in the latent space of the flow, ensuring it can accurately predict the conditional information.
Specifically, we impose a loss during training to ensure a sub-vector $\mathbf{z}_c$ in $\mathbf{m}$ matches the simulation parameters $\mathbf{c}$, where $\mathbf{m} = <..., \mathbf{z}_c>$ is the output of the conditional normalizing flow, as illustrated in~\cref{eq:condflow}. 
Once the model is well-trained, $\mathbf{z}_c$ will accurately approximate $\mathbf{c}$.

During inference, given a simulation outcome $\mathbf{x}$, scientists first utilize encoder $g_e$ to obtain the latent representation $\mathbf{z}_K=g_e(\mathbf{x})$. Then, as illustrated by the black solid arrows in~\cref{fig:nf}, the inverse of the unconditional normalizing flow $f_u$ can predict the simulation parameters $\mathbf{c}$ associated with encoded data as follows:
\vspace{-2pt}
\begin{equation} \label{eq:rev_zc}
\begin{gathered}
<..., \mathbf{z}_c> = f_u^{-1}(g_e(\mathbf{x})), \\
\mathbf{c} = \mathbf{z}_c.
\end{gathered}
\vspace{-5pt}
\end{equation}

In summary, {\sysname} enables an uncertainty-aware bidirectional mapping between simulation parameters and simulation data. 
Given simulation parameters, {\sysname} can produce the corresponding simulation data and quantify the uncertainties in the data generation process. 
In the reverse direction, {\sysname} can predict the simulation parameters from the given data. 
This bidirectional prediction ability is highly beneficial, especially when scientists prefer a robust model for both simulation data generation and simulation parameter prediction, a challenging task for other generative models such as Generative Adversarial Networks (GANs) or Variational Autoencoders (VAEs).


\subsubsection{Loss Functions} \label{sect:lossFunction}
We first train the autoencoder model as discussed in~\cref{sect:ae}, and then train {\sysname} in~\cref{sect:surroflow} and~\cref{sect:reverse_pred}. Our training data are pairs of simulation parameters $\mathbf{c}$ and corresponding simulation data $\mathbf{x}$.

The autoencoder is trained on the simulation data with a Mean Squared Error (MSE) loss between the original data $\mathbf{x}$ and the reconstructed data $\mathbf{x}^{\prime}$:
\begin{equation} \label{eq:loss_ae}
\mathcal{L}_{ae} = \mathbb{E}_{\mathbf{x}}[\|\mathbf{x}-\mathbf{x}^{\prime}\|^{2}_2].
\end{equation}

Training data for {\sysname} are pairs of simulation parameters and latent representations of the corresponding simulation output. Given data $\mathbf{x}$, we first extract its latent representation $\mathbf{z}$ via the trained encoder $g_e$ and utilize $\mathbf{z}$ for {\sysname} training. 
{\sysname} is trained based on a combination of two losses:
\vspace{-2pt}
\begin{equation} \label{eq:loss_total}
\mathcal{L}_{flow} = \mathcal{L}_{f} + \alpha \mathcal{L}_{c}.
\vspace{-1pt}
\end{equation}
The hyperparameter $\alpha$ balances these two losses.
First, we aim to find model parameters that accurately describe the conditional distribution $p(\mathbf{z}\mid \mathbf{c})$, as detailed in~\cref{eq:pz}.
The first loss $\mathcal{L}_{f}$ is the standard log-likelihood loss for normalizing flow training:
\vspace{-2pt}
\begin{equation} \label{eq:loss_flow}
\mathcal{L}_{f} = \mathbb{E}_{\mathbf{z},\mathbf{c}}[-\log p(\mathbf{z} \mid \mathbf{c})].
\end{equation}
This loss minimizes the negative log-likelihood, which is mathematically equivalent to the Kullback-Leibler (KL) divergence between the distribution parameterized by the flow model and the ground truth distribution. 
The second loss $\mathcal{L}_{c}$ minimizes the discrepancy between flow-predicted $\mathbf{z}_c$ and the simulation parameter $\mathbf{c}$. This loss is used to ensure the simulation parameter conditions are accurately captured within the latent space of the normalizing flow:
\vspace{-2pt}
\begin{equation} \label{eq:loss_zc}
\mathcal{L}_{c} = \mathbb{E}_{\mathbf{x},\mathbf{c}} [\| \mathbf{z}_c - \mathbf{c} \|_1].
\vspace{-1pt}
\end{equation}
$\mathcal{L}_{c}$ enables the reverse prediction of simulation parameters for a given simulation output.

\subsection{Genetic Algorithm for Parameter Candidate Generation}\label{sect:ga}

{\sysname} offers accurate data prediction for a given parameter configuration and also the ability to quantify the uncertainties in the data generation process. In this section, we discuss how we leverage the strengths of {\sysname} for efficient simulation parameter space exploration for ensemble data. 
Specifically, we introduce a Genetic Algorithm (GA) for simulation parameter candidate generation, in which {\sysname} helps guide the evolutionary algorithm to identify the simulation parameters that align with scientists' interests. 

\subsubsection{Fitness Function Design} \label{sect:fitness_fun}
Genetic algorithms are widely used to solve complex optimization problems by mimicking the process of natural selection and genetics. 
The key idea behind the genetic algorithm is to operate on a set of candidates in the current generation and improve the population gradually for an optimization goal. This approach is effective for new high-quality candidate discovery.
In this optimization process, a fitness function is crucial in evaluating the ``fitness'' of candidates in each generation. It directly relates to the objective functions of the problem and influences the direction of the evolutionary process. 

Initially, scientists provide several simulation parameters that they might be interested in. The genetic algorithm then identifies new simulation parameters that meet scientists' interests based on quantitative measurements such as data similarity and diversity.
In this case, the simulation outcome for each candidate parameter configuration is required. However, running the complex simulation for all candidates can be prohibitively slow. 
To speed up the optimization process, we utilize {\sysname} to quickly produce the simulation outputs for candidate parameters. 
We also incorporate the uncertainty information produced by the surrogate model {\sysname} into the genetic algorithm. Incorporating uncertainties can increase the robustness and reliability of the parameter recommendation and exploration process. For example, candidates with high uncertainty will have low fitness scores, despite their data being similar to scientists' preferences. 

Our fitness function considers the similarity, diversity, and uncertainty of candidate solutions (i.e., parameter sets). 
For a candidate simulation parameter configuration $\mathbf{c}$, we utilize {\sysname} to predict data $\bar{\mathbf{x}}$ and associated uncertainty $\mathbf{x}_{var}$, as described in~\cref{alg:uq}. 
Similarly, for a collection of simulation parameters $C_{usr}$ selected by scientists, {\sysname} predicts the data list $X_{usr}$. For the set of selected parameters $C_{usr}$, scientists will assign preference scores $S_{usr}$ $\in [-1, 1]$ to each of them. A higher value indicates scientists are more interested in this parameter configuration. 
Our fitness function is defined as
\vspace{-2pt}
\begin{equation} \label{eq:ga_fitness}
\begin{gathered}
F(\mathbf{c}, C_{usr}, S_{usr}) \\= w_1 \cdot \text{sim}(\bar{\mathbf{x}}, X_{usr}, S_{usr}) + w_2 \cdot \text{div}(\mathbf{c}, C_{usr}) + w_3 \cdot \text{unc}(\mathbf{x}_{var}),
\end{gathered}
\end{equation}
where $w_1$, $w_2$, and $w_3$ are weights in range $[-1, 1]$ to balance these metrics. By adjusting these weights appropriately, the fitness function in~\cref{eq:ga_fitness} can favor solutions with different objectives. 
For example, find simulation parameters that are diverse but have simulation outcomes similar to the user's preferences, while also encouraging low uncertainty in predicted data. 
This function can also be customized to identify simulation parameters with high uncertainty, helping training data augmentation for {\sysname}.
The three components of the fitness function are:

\begin{enumerate}
\item \textbf{Similarity Score} measures how closely the predicted data $\bar{\mathbf{x}}$ of a candidate parameter $\mathbf{c}$, matches the user preferred outcomes $X_{usr}$:
\vspace{-2pt}
\begin{equation} \label{eq:fitness_sim}
\text{sim}(\bar{\mathbf{x}}, X_{usr}, S_{usr}) = \sum_{\substack{\bar{\mathbf{x}}_{usr} \in X_{usr}, s_{usr} \in S_{usr}}} \frac{1}{\text{distance}(\bar{\mathbf{x}}, \bar{\mathbf{x}}_{usr})} \cdot s_{usr}.
\end{equation}
Similar to content-based filtering widely used in modern recommendation systems, this metric recommends simulation parameters that have similar outputs to the user's preferences. The inverse distance is used to measure similarity. Candidates whose data are similar to the user's selection with high user-specified preference scores in $S_{usr}$ will have high similarity scores.

\item \textbf{Diversity Score} evaluates how distinct $\mathbf{c}$ is compared to the user preferred set of parameters $C_{usr}$:
\vspace{-2pt}
\begin{equation} \label{eq:fitness_div}
\text{div}(\mathbf{c}, C_{usr}) = \sum_{\substack{\mathbf{c}_{usr} \in N_k(\mathbf{c})}} \text{distance}(\mathbf{c}, \mathbf{c}_{usr}).
\vspace{-1pt}
\end{equation}
This metric encourages the variety among the recommended simulation parameters. Formally, the diversity is measured as the total distance to $k$ nearest neighbors of $\mathbf{c}$ in the user's selection set $C_{usr}$, represented as $N_k(\mathbf{c})$. We set $k=5$.

\item \textbf{Uncertainty Score} quantifies the variance in the {\sysname}'s output $\bar{\mathbf{x}}$: 
\vspace{-2pt}
\begin{equation} \label{eq:fitness_unc}
\text{unc}(\mathbf{x}_{var}) = \frac{1}{n} \sum_{i=1}^{n} {(\mathbf{x}_{var})}_i.
\end{equation}
It is measured as the mean of the estimated uncertainty field, where ${(\mathbf{x}_{var})}_i$ is the uncertainty value at the $i$-th index of $\mathbf{x}_{var}$ and $n$ is the dimensionality of $\mathbf{x}_{var}$. High variance indicates low confidence in the model's output. 

\end{enumerate}

To speed up the similarity and uncertainty computation during optimization, the calculation is based on the latent representations of data generated by {\sysname}, which can be done much more efficiently than using the reconstructed data generated by the decoder.
This avoids the costly and unnecessary full reconstruction of data for similarity comparison. Hence, $\text{distance}(\bar{\mathbf{x}}, \bar{\mathbf{x}}_{usr})$ in~\cref{eq:fitness_sim} is evaluated using the inverse of cosine similarity between latent representations, while $\text{distance}(\mathbf{c}, \mathbf{c}_{usr})$ in~\cref{eq:fitness_div} is based on L1 differences between candidate and user-selected parameters. 
The associated uncertainty is assessed through the variance of these latent representations. 

\vspace{-4pt}
\begin{algorithm}
\caption{Simulation parameter candidate generation using GA}\label{alg:ga}
\begin{algorithmic}[1]
\State \textbf{Input:} Population size $p$, maximum generations $g$, user-selected parameters $C_{usr}$, user preference scores $S_{usr}$, mutation rate $r_m$
\State \textbf{Output:} Optimal set of simulation parameters $C$
\Procedure{OptimizeParams}{$p$, $g$, $C_{usr}$, $S_{usr}$, $r_m$}
    \State $C \gets$ randomly initialized population with $p$ samples
    \For{$i = 1$ to $g$}
        \State $fitness \gets F(C, C_{usr}, S_{usr})$
        \State $parents \gets \text{select}(C, fitness)$
        \State $C \gets \text{crossover}(parents)$
        \State $C \gets \text{mutate}(C, r_m)$
    \EndFor
    \State \textbf{return} $C$
\EndProcedure
\end{algorithmic}
\end{algorithm}
\vspace{-6pt}

\subsubsection{Candidates Optimization Process}
As outlined in~\cref{alg:ga}, the genetic algorithm begins with a set of randomly initialized candidates for the first generation. 
Subsequent generations proceed as follows: all current candidates are evaluated using the fitness function in~\cref{eq:ga_fitness}, where higher scores indicate better solutions. 
Then, selection is performed to choose parent candidates for reproduction. Samples with higher fitness scores will have a high probability of being chosen. 
After selection, crossover is applied, i.e., candidate parameter vectors are divided into chunks and recombined to create new offspring. This crossover ensures that the offspring inherits characteristics from both parents.
To ensure diversity and prevent local optima, the mutation is then performed at a rate of $r_m$ by adding Gaussian noise to the candidate parameter. The Gaussian noise has a mean of zero and a standard deviation of $r_\sigma$. We set $r_m = 0.2$ and $r_\sigma = 0.1$. This mutation introduces controlled variability which is essential for effective exploration of the simulation parameter space.
These new candidates form the subsequent generation. The optimization process terminates when it reaches the maximum number of generations. After generations of optimization, we can identify simulation parameters that align with scientists' preferences.

\subsection{Visual Interface for Parameter Space Exploration}

We integrate the trained {\sysname} with the genetic algorithm as the backend and develop a visual interface to assist exploration of the vast parameter space. 
This combination improves the efficiency of the exploration process by leveraging scientists' knowledge to identify simulation parameters of interest.

\begin{figure}[htp]
    \centering
    \includegraphics[width=0.9\columnwidth]{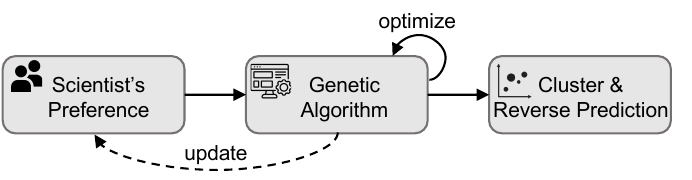}
    \vspace{-5pt}
    \caption{An overview of our visual system for efficient user-guided parameter recommendation and exploration.}
    \label{fig:system_overview}
    \vspace{-3pt}
\end{figure}

The visual interface facilitates the parameter exploration process in three ways. 
First, it displays volume rendering images of the {\sysname}'s prediction results and the quantified uncertainties for user-selected simulation parameters, enabling informed decision-making. 
Second, the visual system enables scientists to assign preference scores to candidates after carefully inspecting the visualization results. 
Then, the genetic algorithm in the backend computes fitness scores for candidate simulation parameters according to scientists' preferences, guiding the optimization toward scientists' preferred directions. 
Third, the interface provides an overview of the genetic algorithm's optimization process across generations, offering insights into its progression.
An overview of the visual exploration and optimization process is shown in~\cref{fig:system_overview}. It contains three major steps.

\vspace{-12pt}

\begin{figure}[htp]
    \centering
    \includegraphics[width=\columnwidth]{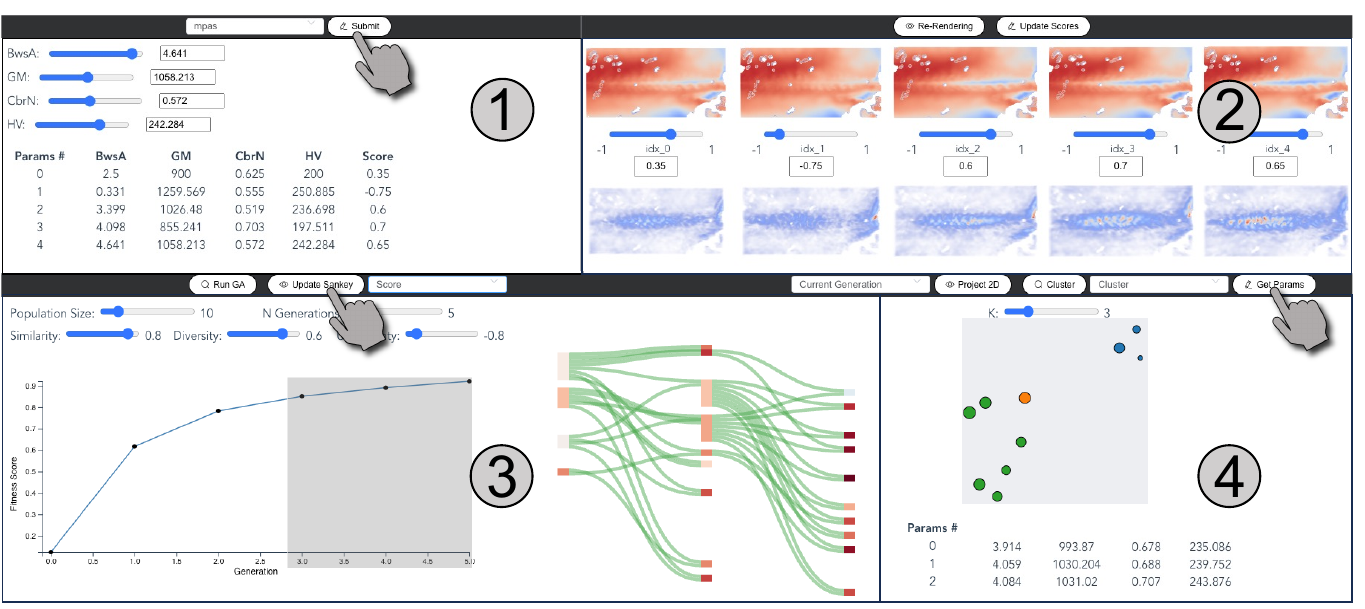}
    \vspace{-12pt}
    \caption{Visual interface for parameter recommendation and exploration. (1) Simulation parameter selection view to select scientist-interested parameters. (2) The result visualization view to assign scientists' preference scores for each parameter configuration. (3) Genetic algorithm view to interactive control and visualize the optimization process. (4) Projection view to visualize the clustering result and the predicted parameters. }
    \label{fig:interface_eg}
    \vspace{-8pt}
\end{figure}

\textbf{Step 1: Specify the scientist's preference.}
In the simulation parameter selection view of the visual interface (\cref{fig:interface_eg}~(1)), scientists choose multiple simulation parameters to define their interests. The trained surrogate model {\sysname} in the backend can efficiently generate data for the selected simulation parameters. The reconstruction and uncertainty quantification results of the selections are visualized in \cref{fig:interface_eg}~(2). 
After carefully analyzing the visualization results, scientists can assign preference scores to these selected parameters. A higher score indicates a higher interest. The user-selected parameters and corresponding preference scores are recorded in the table in \cref{fig:interface_eg}~(1). 

\textbf{Step 2: Optimize through the genetic algorithm.}
The second step is to optimize the randomly initialized candidate parameters through the genetic algorithm running in the backend of the visual system. The optimization process is controlled and visualized in \cref{fig:interface_eg}~(3). 
In this view, scientists can interactively assign different weights for similarity, diversity, and uncertainty for different optimization objectives, as discussed in~\cref{sect:ga}.
The initialized simulation parameters will then evolve through selection, crossover, and mutation based on the calculated fitness scores in each generation. 
The average fitness scores over generations are displayed in the line graph, where the x-axis represents the generation index and the y-axis represents the fitness score value. 
By brushing on the line graph, the Sankey diagram will be updated. Nodes in the Sankey represent candidate parameters and links depict the parent-child relationship. Node color is set via a drop-down menu, allowing color by similarity, diversity, or uncertainty. Each step in the Sankey diagram represents one generation. Hovering over a link highlights its ancestors and descendants from the first to the last generation within the brushed area.
The iterative optimization ultimately presents scientists with simulation parameters that closely match their interests. Moreover, if scientists identify highly interesting new parameters during optimization, they can make further selections by clicking on nodes in the Sankey diagram to refine their selected parameter set for future optimization rounds.

\textbf{Step 3: Cluster and reverse prediction of representative simulation parameters.}
Given the potentially large set of recommended simulation parameters, running all recommendations through {\sysname} and visually inspecting all rendering results is impractical. To address this, we extract representative simulation parameters as our final recommendation. 
For all candidate simulation parameters in the last generation of the genetic algorithm, we first retrieve latent representations for data samples corresponding to these parameters. K-means clustering is then applied to take cluster centers as representative latent representations. We utilize the reverse prediction ability of~{\sysname} to determine the simulation parameters for these cluster centers. 
We also project all latent representations onto a 2D plane using t-SNE (t-distributed Stochastic Neighbor Embedding) and visualize the clustering and reverse prediction results in \cref{fig:interface_eg}~(4).

\vspace{-8pt}

\section{Results} \label{sect:Results}
We evaluate {\sysname}'s ability as a surrogate model for data generation, uncertainty quantification, and reverse prediction. We also conduct a case study for simulation parameter recommendation and exploration with our visual interface.

\subsection{Dataset and Implementation}
The proposed model {\sysname} is evaluated using two scientific ensemble simulation datasets in ~\cref{tab:dataset}, which also includes the simulation parameter ranges for both datasets.

\textbf{MPAS-Ocean}~\cite{ringler2013mpas} dataset contains the simulation results from the MPAS-Ocean model for ocean system simulation, developed by the Los Alamos National Laboratory. 
According to domain scientists' interests, we study four input parameters: Bulk wind stress Amplification ($BwsA$), Gent-McWilliams Mesoscale eddy transport parameterization ($GM$), Critical Bulk Richardson Number ($CbrN$), and Horizontal Viscosity ($HV$). 
During training, 128 parameter settings and corresponding data samples with a resolution of 
$192 \times 96 \times 12$ are used. For testing, 20 parameter settings are utilized.

\textbf{Nyx}~\cite{almgren2013nyx} is a cosmological simulation dataset from compressible cosmological hydrodynamics simulations by Lawrence Berkeley National Laboratory. 
Based on domain scientists' suggestions, we study three input parameters: the Omega Matter parameter ($OmM$) representing the total matter density, the Omega Baryon parameter ($OmB$) representing the density of baryonic matter, and the Hubble parameter ($h$) quantifying expansion rate of the universe. We utilize $70$ parameter configurations for training and 20 for testing. The log density field of resolution $128 \times 128 \times 128$ is used for experiments.

{\sysname} is developed based on  PyTorch$\footnote{https://pytorch.org}$ and is trained on a single NVIDIA A100 GPU with a learning rate of $10^{-4}$. 
The visual system is implemented with Vue.js$\footnote{https://vuejs.org/}$ for the frontend and Flask$\footnote{https://flask.palletsprojects.com/}$ for the backend server. VTK.js$\footnote{https://kitware.github.io/vtk-js/}$ is used for volume rendering of data. 

\vspace{-4pt}
\begin{table}[!ht]
\small
\caption{Simulation dataset name and simulation parameter range.}
\vspace{-8pt}
\centering
 \begin{tabular}{c|c} 
 Dataset & Simulation Parameter Range\\ [0.5ex] 
 \hline
 MPAS-Ocean & \makecell{$BwsA \in [0.0, 5.0]$, $GM \in [600.0, 1500.0]$,\\ $CbrN \in [0.25, 1.00]$, $HV \in [100.0, 300.0]$} \\
 \hline 
 Nyx & \makecell{$OmM \in [0.12, 0.55]$, $OmB \in [0.0215, 0.0235]$, $h \in [0.55, 0.85]$} \\
 \end{tabular}\label{tab:dataset}
\vspace{-10pt}
\end{table}

\vspace{-6pt}
\begin{table}[!ht]
\small
\caption{Model size, training time, and test time for each dataset.}
\vspace{-8pt}
\centering
 \begin{tabular}{c|c|c|c|c} 
 Dataset & Model Name & Model size & Training time & Test time\\ [0.5ex] 
 \hline
 \multirow{2}*{MPAS-Ocean} & VDL & 0.63 GB & 139.8h & 21.0s \\
 ~ & {SurroFlow(our)} & 47.93MB  & 11.5h & 0.8s \\
 \hline 
 \multirow{2}*{Nyx} & VDL & 1.98 GB & 82.7h & 9.2s \\
 ~ & {SurroFlow(our)} & 42.93MB & 47.5h & 0.9s \\
 \end{tabular}\label{tab:model}
\vspace{-10pt}
\end{table}

\subsection{Quantitative Evaluation} \label{sect:QuantiEval}
In this section, we quantitatively evaluate {\sysname}'s performance in terms of surrogate modeling and reverse prediction. 

\subsubsection{Surrogate Prediction} 

We use two metrics to assess the data generation quality of {\sysname} for a given simulation parameter input. 
For data-level evaluation, we employ peak signal-to-noise ratio (PSNR) to quantify the voxel-level differences between surrogate-generated data and actual simulation results. 
For image-level evaluation, we utilize the structural similarity index measure (SSIM) to evaluate the quality of volume rendering images from generated data against the simulation outcomes. Higher values of PSNR and SSIM indicate better quality. 

As discussed in~\ref{sect:surrogate}, {\sysname} contains two components: an autoencoder (AE) for data reduction and a flow-based model for distribution modeling. The flow-based model is trained on the AE's latent space. The quality of the AE's reconstructions is critical as it directly impacts the effectiveness of the flow-based modeling. Therefore, we first evaluate AE's reconstruction quality using the above metrics. 
As shown in~\cref{tab:eval}, the AE model achieves high PSNR and SSIM scores for test data of all datasets, indicating minimal information loss in latent representations. This enables the {\sysname} to train on latent representations, reducing computational costs compared to using raw data.

Then we compare {\sysname}'s prediction ability with one state-of-the-art baseline, i.e., VDL-Surrogate~\cite{shi2022vdl} (shorted for VDL), a view-dependent surrogate model that utilizes latent representations from selected viewpoints for fast training and interpolation during inference. 
We use the original implementation of VDL. The model sizes and training time for VDL and {\sysname} are detailed in~\cref{tab:model}.
The quantitative results in~\cref{tab:eval} show that {\sysname} can achieve comparable performance with the baseline. {\sysname} performs slightly better on the MPAS-Ocean dataset. For the Nyx dataset, since VDL utilizes an importance-driven loss to ensure high-density values are preserved, it performs better than {\sysname}. 
However, {\sysname} offers additional benefits including quantifying uncertainties in the surrogate model and predicting simulation parameters for given simulation outcomes.

\begin{table}[!ht]
\small
\vspace{-4pt}
\caption{PSNR and SSIM for AE, {\sysname} and baseline's results. }
\vspace{-8pt}
\centering
 \begin{tabular}{c|c|c|c}
  Data & Method & $\uparrow$PSNR & $\uparrow$SSIM  \\ [0.5ex] 
  \hline
  \multirow{3}*{MPAS-Ocean} & VDL & 45.8223 & 0.9842 \\
  ~ & AE (our) & 50.1931 & 0.9963 \\
  ~ & {\sysname}(our) & 46.6852 & 0.9950 \\
  \cline{1-4}
  \multirow{3}*{Nyx} & VDL & 33.6638 & 0.9160 \\
  ~ & {AE} (our) & 36.6843 & 0.8865 \\
  ~ & {\sysname}(our) & 30.9263 & 0.8304 \\
  \cline{1-4}
 \end{tabular} \label{tab:eval}
\vspace{-6pt}
\end{table}

\subsubsection{Reverse Prediction} 
Different from other surrogate models, {\sysname} has the bidirectional prediction ability.  
In this section, we measure the ability of {\sysname} to predict simulation parameters for a given simulation outcome.  

To assess the accuracy of predicted simulation parameters against the ground truth, we employ mean absolute error (MAE) and cosine similarity as our evaluation metrics. MAE is lower the better and cosine similarity is higher the better. 
Given that the parameters span diverse ranges, before we evaluate them with the metrics, we apply min-max normalization to every dimension of the predicted parameter vector based on the range detailed in~\cref{tab:dataset}. This evaluation for simulation parameter prediction is conducted on 20 randomly selected simulation data for each dataset.

\Cref{tab:param_mae} displays the average quantitative evaluation results on all test data of each dataset. Given the low MAE and high similarity of our reverse prediction results to the ground truth simulation parameters, it's evident that our model achieves significant accuracy in reconstructing the original simulation conditions.
In~\cref{tab:params}, we also represent some examples of the parameter prediction results. Comparing the predicted parameters with the ground truth, we observe a close alignment, demonstrating the effectiveness of our model in simulation parameter prediction through the reverse direction of the surrogate model.

\begin{table}[!ht]
\small
\vspace{-4pt}
\caption{MAE and cosine similarity for {\sysname}'s parameter predictions. }
\vspace{-6pt}
\centering
 \begin{tabular}{c|c|c}
  Data & $\downarrow$MAE & $\uparrow$Cosine \\ [0.5ex] 
  \hline
  {MPAS-Ocean} & 0.1958 & 0.8624 \\
  \cline{1-3}
  {Nyx} & 0.0050 & 0.9999 \\ 
  \cline{1-3}
 \end{tabular} \label{tab:param_mae}
\vspace{-9pt}
\end{table}

\begin{table}[!ht]
\small
\vspace{-5pt}
\caption{Examples of predicted and ground truth simulation parameters}
\vspace{-6pt}
\centering
 \begin{tabular}{c|c}
  Data & Parameters \\ [0.5ex] 
  \hline
  \multirow{6}*{MPAS-Ocean} & \makecell{GT: [3.4908, 1106.3879, 0.4773, 173.4463] \\ Pred: [3.5231, 816.6441, 0.4641, 222.8455]} \\
  \cline{2-2}
  ~ & \makecell{GT: [1.7097, 522.8835, 0.2817, 207.0228] \\ Pred: [1.7297, 634.0682, 0.2850, 211.4337]} \\
  \cline{2-2}
  ~ & \makecell{GT: [2.7745, 957.3225, 0.8250, 148.2813] \\ Pred: [2.7160, 1184.3029, 0.8356, 136.5965]} \\
  \cline{1-2}
  \multirow{6}*{Nyx} & \makecell{GT: [0.1389, 0.0224, 0.8005] \\ Pred: [0.1405, 0.0271, 0.7949]} \\
  \cline{2-2}
  ~ & \makecell{GT: [0.1518, 0.0216, 0.6606] \\ Pred: [0.1545, 0.0231, 0.6633]} \\
  \cline{2-2}
  ~ & \makecell{GT: [0.1424,0.0228, 0.5522] \\ Pred: [0.1307, 0.0234, 0.5533]} \\
  \cline{1-2}
 \end{tabular} \label{tab:params}
\vspace{-5pt}
\end{table}

\vspace{-6pt}
\begin{figure}[htp]
    \centering
    \includegraphics[width=0.95\columnwidth]{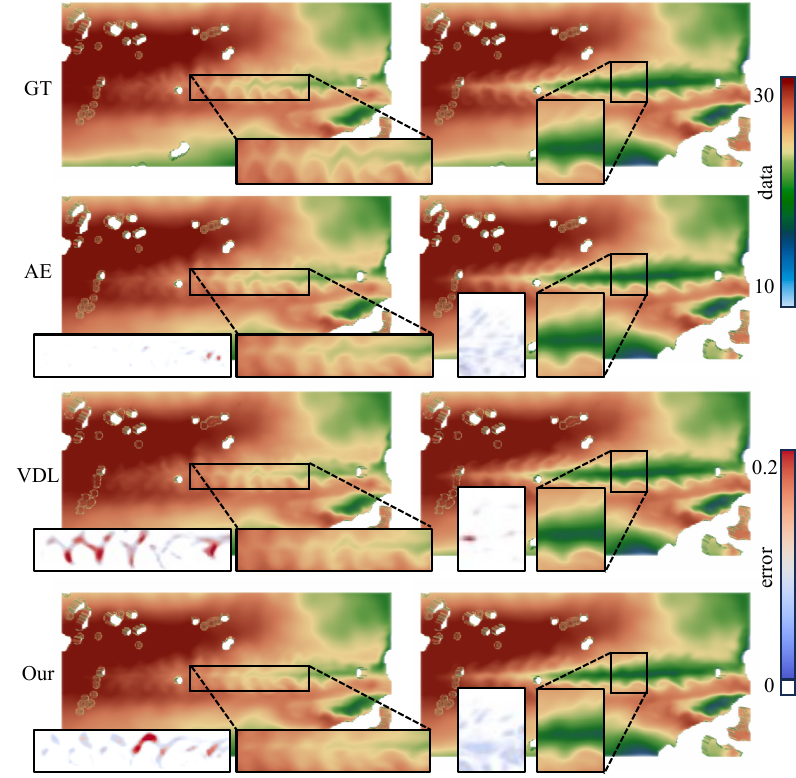}
    \vspace{-5pt}
    \caption{Volume rendering of MPAS-Ocean data for ground truth, predicted data, and squared error maps for AE, baseline VDL, and {\sysname}. Each column shows results from one simulation parameter configuration.
    }
    \label{fig:maps}
    \vspace{-5pt}
\end{figure}

\begin{figure}[ht]
    \centering
    \includegraphics[width=0.9\columnwidth]{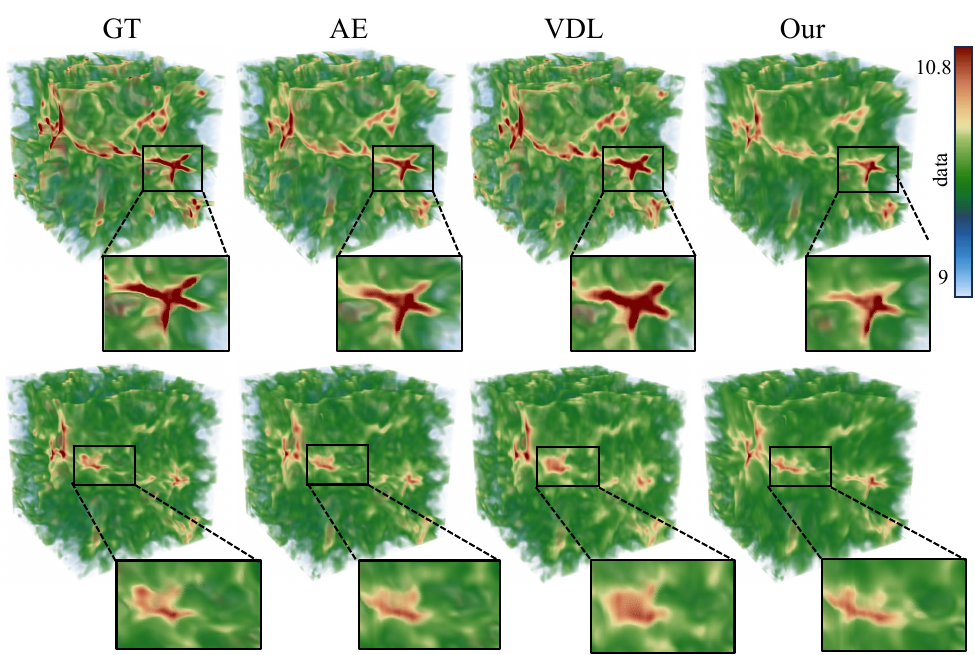}
    \vspace{-5pt}
    \caption{Volume rendering of Nyx data for ground truth, predicted data from AE, baseline VDL, and {\sysname}. Each row shows results from one simulation parameter configuration.}
    \label{fig:nyx}
    \vspace{-8pt}
\end{figure}

\subsection{Qualitative Evaluation}\label{sect:QualiEval}
In this section, we qualitatively compare the {\sysname}'s results with the baseline model, VDL-Surrogate, via the volume rendering images. The volume rendering settings are kept the same for the same dataset. 

\Cref{fig:maps} compares the volume rendering images of the ground truth data with the predicted ones on the MPAS-Ocean dataset. There are three neural network models in comparison, including an autoencoder (AE) for data reduction and two surrogate models, i.e., the baseline VDL and the proposed model {\sysname}. The two columns show results from two different simulation parameter configurations. 
In each result, we provide a zoom-in view at the bottom of volume rendering images for better comparison across models. The zoom-in region is near the region of interest for domain scientists called the Pacific Cold Tongue where the temperatures drop to about $24^\circ\text{C}$ in this area. On the left side of the zoom-in figures, we also show the corresponding squared error map where we set the white color to low error regions. 
By comparing the rendering results of all models with the ground truth, we found that overall the AE used for latent representation extraction achieves the best results, with low errors and a clearer boundary of the feature in the zoom-in region. 
Comparing {\sysname} with the baseline model VDL, we found {\sysname} on average has a low error, and VDL can have higher errors in some feature boundaries.

\Cref{fig:nyx} shows volume rendering images of Nyx data, comparing the ground truth with images generated by the baseline model VDL, the autoencoder (AE), and {\sysname}. Each row corresponds to the results from one simulation parameter setting. 
The AE model exhibits superior reconstruction quality due to its relatively straightforward task. Although both the baseline model VDL and {\sysname} successfully capture crucial features, VDL shows more clear high-density details than {\sysname}. However, VDL tends to exaggerate high-density outputs, as shown in the zoomed-in views for both rows. 
Regions expected to have low-density values now have relatively high values. The high-density feature preservation ability of VDL comes from their adoption of importance-driven loss, which may amplify the density values into a higher range. The proposed {\sysname} does not incorporate such important information, so the results are less biased, and with uncertainty quantification ability, {\sysname} is more reliable.  

In summary, {\sysname} offers performance on par with the baseline model VDL when serving as a surrogate model. However, {\sysname} provides features like uncertainty quantification in the surrogate modeling process and the capability for reverse prediction of simulation parameters for the given simulation outcomes, functions that are not available for the baseline.


\subsection{Uncertainty Quantification}
Due to the explicit distribution modeling, {\sysname} captures the variations of predicted simulation outputs for the conditional input simulation parameters. 
In this section, we assess the uncertainties in the surrogate model {\sysname} using the method outlined in~\cref{alg:uq}. The number of samples for uncertainty quantification $N$ is set to 20.

\begin{figure}[htp]
    \centering
    \includegraphics[width=\columnwidth]{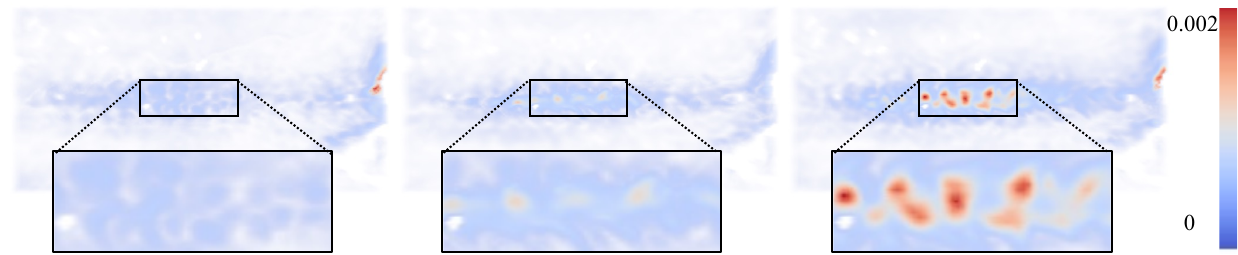}
    \vspace{-14pt}
    \caption{Volume rendering of the estimated uncertainty field for MPAS-Ocean dataset. From left to right, parameters $GM$, $CbrN$, and $HV$ are the same, parameter $BwsA$ is increasing from $0.5$, $2.5$, to $4.5$.}
    \label{fig:maps_uq_vr}
    \vspace{-8pt}
\end{figure}

In~\cref{fig:maps_uq_vr}, we show the volume rendering images of the estimated uncertainty field for the MPAS-Ocean dataset. These images are generated from three different simulation parameter configurations. We set parameters $GM$, $CbrN$, and $HV$ to $900$, $0.625$, and $200$, and explore the uncertainty introduced by the surrogate model {\sysname} when we change the parameter $BwsA$. The $BwsA$ values are set to $0.5$, $2.5$, and $4.5$ from left to right, respectively. As the value of $BwsA$ increases, the uncertainty values in the eastern Pacific region also rise, as detailed in the zoom-in images in~\cref{fig:maps_uq_vr}. Other regions maintain relatively low uncertainty levels. 
The uncertainty comes from the model's sensitivity to $BwsA$ values, highlighting areas where {\sysname} can be less reliable. Despite this, the overall unreliability range remains at a lower level, underscoring the model's overall robustness.

\subsection{Case study: Simulation Parameter Space Exploration}\label{sect:casestudy}

This section demonstrates the effectiveness of our visual system for efficient simulation parameter recommendation and exploration. 

\vspace{-9pt}
\begin{figure}[htp]
    \centering
    \includegraphics[width=\columnwidth]{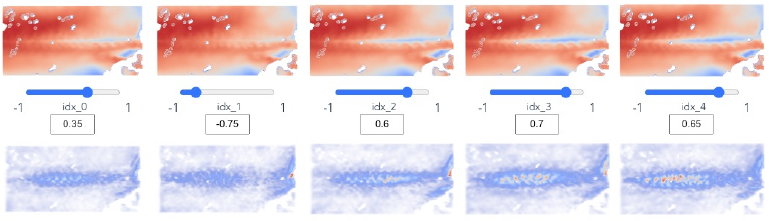}
    \vspace{-16pt}
    \caption{Scientists' selections and corresponding preference scores.}
    \label{fig:maps_selection}
    \vspace{-5pt}
\end{figure}

The case study is to explore the simulation parameter space for the MPAS-Ocean dataset, which includes four simulation parameters detailed in~\cref{tab:dataset}. 
The case study involves a scientist with over 15 years of experience in ocean science simulation analysis and model development.
The scientist aims to analyze the cold tongue in the eastern equatorial Pacific Ocean. The cold tongue phenomenon is caused by the upwelling of deep, cold waters to the sea surface, forming a strip of cold water stretching along the equator. 
In the exploration process, with the interactive visual system, the scientist starts by selecting several simulation parameters randomly within their valid ranges. This provides a starting point for exploration, after which domain knowledge is applied to refine and guide the parameter exploration based on the scientist's expertise and the visualization results.
After selection, {\sysname} running in the backend will predict the corresponding simulation outputs for the input simulation parameters. 
Based on the volume-rendered images of {\sysname}'s outputs (as shown in \cref{fig:maps_selection}), the scientist interactively assigns high preference scores to those with a relatively lower temperature in the east Pacific region, a feature he is interested in. The visual system uses a ``Cool to Warm'' colormap for the MPAS-Ocean dataset, allowing for intuitive temperature visualization. Parameters predicting higher temperatures in the target area are assigned with low preference scores to refine the focus on relevant simulations.

The weights in the genetic algorithm's fitness function were determined iteratively, guided by the scientist's domain knowledge and practical evaluation. Starting with initial guesses based on the optimization goal, the scientist gets the desired weights for similarity, diversity, and uncertainty ($0.8$, $0.6$, and $-0.8$, respectively) through repeated testing.
This configuration aims to discover new parameters that closely match the preferred simulation outcomes while ensuring diversity and reliability. The negative weight for uncertainty is strategically chosen to penalize parameters with high uncertainty. 
The quantified uncertainty stems from surrogate model approximation errors. Focusing on low-uncertainty parameters ensures reliable predictions for fitness function computation and exploration. However, high-uncertainty regions can also offer valuable insights, highlighting areas that need improvement and guiding future simulations and data augmentation to enhance model quality and reliability.

\begin{figure}[htp]
    \centering
    \includegraphics[width=\columnwidth]{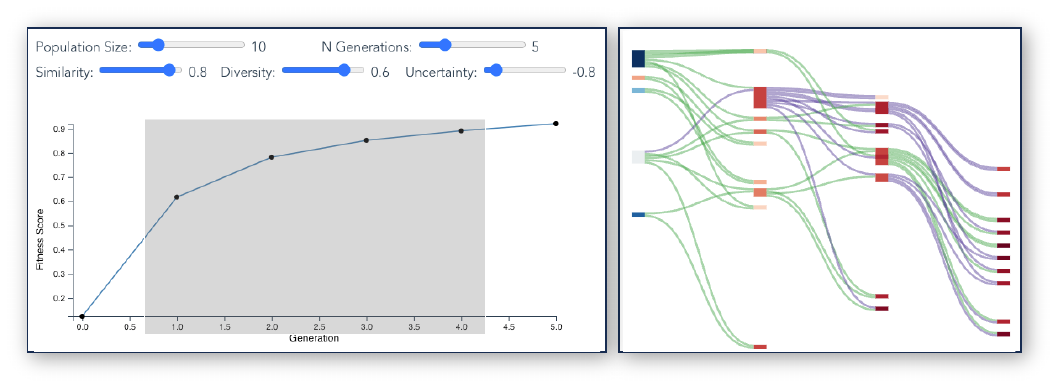}
    \vspace{-18pt}
    \caption{Left: fitness scores over iterations of the genetic algorithm optimization. Right: The Sankey diagram corresponds to the brushed region on the line plot. Nodes are color-encoded by the fitness values.}
    \label{fig:maps_fitness}
    \vspace{-6pt}
\end{figure}

With the genetic algorithm optimized for $5$ generations, the fitness scores plateau. 
As illustrated in the line chart in~\cref{fig:maps_fitness}, there is a clear increasing trend in the fitness score over generations. By brushing from generation $1$ to $5$ on the line chart, the Sankey diagram will be updated, illustrating the evolution of candidates across these generations. 
Nodes in the Sankey diagram are color-coded based on the fitness scores, with darker colors indicating higher fitness scores. The overall fitness scores increase from left to right across generations. 
Nodes with higher fitness scores are more likely to be chosen as parents for the next generation. Nodes with larger lengths are those with a higher number of offspring. By hovering over a link in the Sankey diagram, the connected paths are highlighted in purple, tracing to the node's ancestors and descendants. 
Scientists can interact with the Sankey diagram by clicking on nodes to include the associated parameter configurations in the preferred selections and assign preference scores.

\vspace{-8pt}
\begin{figure}[htp]
    \centering
    \includegraphics[width=\columnwidth]{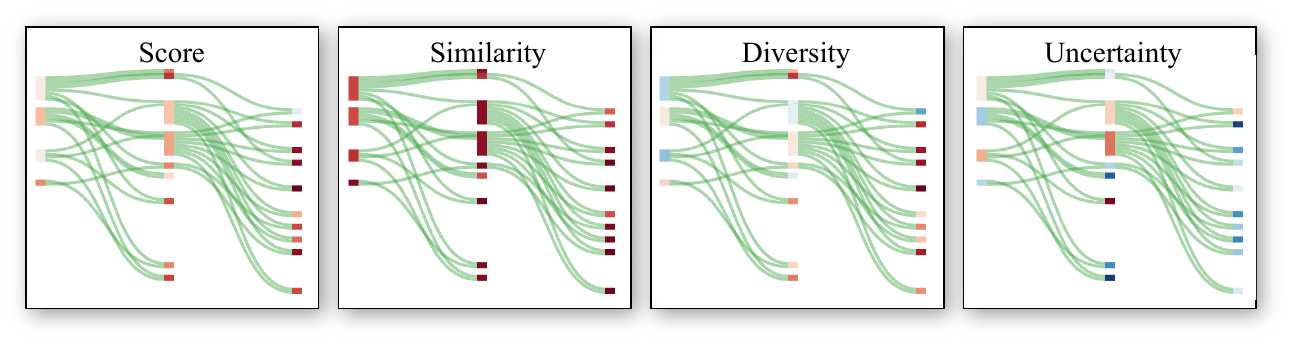}
    \vspace{-18pt}
    \caption{Sankey diagrams illustrate the evolution of fitness, similarity, diversity, and uncertainty scores across generations. These scores align with the scientist's exploration goal of enhancing similarity and diversity while reducing uncertainty.}
    \label{fig:sankey}
    \vspace{-2pt}
\end{figure}

By altering the colors of the nodes in the Sankey diagram, scientists can gain insights into the simulation parameter optimization process. As depicted in~\cref{fig:sankey}, for the brushed region from generation $3$ to $5$, nodes in the Sankey diagram are color-coded according to each candidate's fitness score, similarity, diversity, and uncertainty, respectively. 
It is clear that, as the evolutionary algorithm proceeds, the similarity scores as well as the diversity scores for candidates are increased across generations. Furthermore, given the algorithm's design to minimize uncertainty (indicated by a negative weight on uncertainty), there is a clear trend toward reduced uncertainty. The Sankey results demonstrate the parameters optimization process aligns with the predefined scientists' objectives. 

\begin{figure}[htp]
    \centering
    \includegraphics[width=0.6\columnwidth]{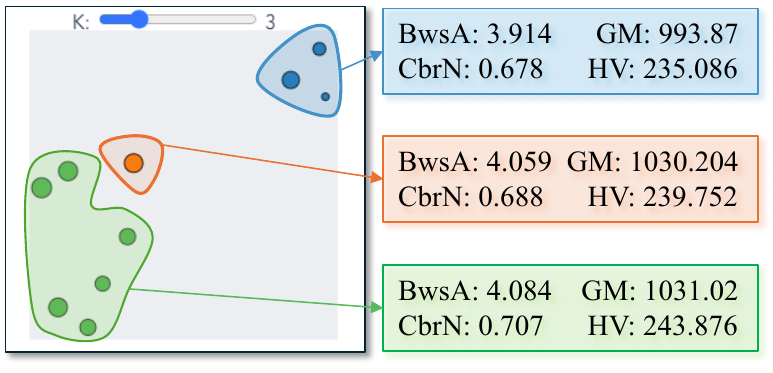}
    \vspace{-8pt}
    \caption{Clustering and prediction for parameter recommendation.}
    \label{fig:sankey_cluster}
    \vspace{-6pt}
\end{figure}

Given that each generation typically contains a lot of candidates, visualizing and analyzing results for all candidates may have duplicates and be inefficient. To address this, our visual system employs the K-means clustering algorithm to group the optimized candidates into $K$ clusters based on their latent representations, as shown in~\cref{fig:sankey_cluster}. 
Then, the system leverages {\sysname}'s reverse prediction ability to derive $K$ representative parameter configurations from the $K$ cluster centers. We set $K=3$ in the experiment. As shown in~\cref{fig:sankey_cluster}, the recommended parameters settings are (1) $BwsA=3.914, GM=993.87, CbrN=0.678, HV=235.086$, (2) $BwsA=4.059, GM=1030.204, CbrN=0.688, HV=239.752$, and (3) $BwsA=4.084, GM=1031.02, CbrN=0.707, HV=243.876$.
Through simulation runs, the scientist verifies that these identified configurations produce simulation outcomes that match the expected behavior of the cold tongue, a phenomenon he is interested in within the MPAS-Ocean dataset. 
Analyzing the cold tongue phenomenon is crucial because it is closely linked to significant climate patterns like El Ni\~{n}o and La Ni\~{n}a, which greatly impact global weather and climate variability. Accurate modeling of the cold tongue is essential for climate simulations. Our system's recommended parameters, such as higher $BwsA$ values enhancing the cold tongue effect, provide a deep understanding of the contributing factors. Additionally, the system can identify less obvious configurations that influence the cold tongue, potentially guiding future research.
This case study shows the {\sysname} and genetic algorithm-assisted visual system helps efficient simulation parameter exploration and recommendation. 

\vspace{-8pt}


\section{Discussion and Future work} 
Compared to existing surrogate models, the proposed {\sysname} has extra benefits such as uncertainty quantification and bidirectional prediction. 
Using {\sysname} and the genetic algorithm as the backend, we have demonstrated the effectiveness of the visual system for parameter space exploration driven by scientists' preferences. 
However, there are still several future directions we can explore. 

First, although the normalizing flow excels at learning complex distributions and has shown outstanding performance in {\sysname}, its learning ability is constrained by the invertible requirement. Other more powerful generative models, such as denoising diffusion models~\cite{ho2020denoising}, can serve as potential alternatives for our task. However, they lack the bidirectional prediction ability. 
Second, in our case study, scientists prioritize data samples with low uncertainty to ensure reliability. However, samples with high uncertainty are also valuable. They reflect the areas where the network does not model well. By incorporating those highly uncertain data samples into the training dataset, we can enhance the robustness of the surrogate model. 
Third, applications for the scientist-guided simulation parameter recommendation can be broader. The proposed simulation parameter exploration system can be useful for a variety of real-world problems. In the future, we would like to explore the possibility of applying {\sysname} across different domains for various tasks such as optimizing engineering design and scientific discovery. 


\section{Conclusion} 
In this paper, we introduce {\sysname}, a novel surrogate model for efficient simulation parameter recommendation and exploration. 
{\sysname} effectively captures the conditional distribution of simulation outcomes conditioned on the simulation parameters, facilitating efficient data generation for different simulation parameters. 
The complex conditional distribution is modeled by applying a series of invertible and learnable transformations on a base Gaussian distribution. The variance in the Gaussian distribution reflects the uncertainty in the surrogate modeling process. Remarkably, these transformations also enable reverse computation, allowing for the prediction of simulation parameters from simulation data.
We integrate {\sysname} with a genetic algorithm and utilize them as the backend for a visual interface, where scientists can interactively specify their optimization objectives through the visual system. After that, the optimization process starts automatically for scientists' preference-guided parameter exploration. 
Our quantitative and qualitative results demonstrate {\sysname}'s ability to predict high-fidelity simulation outputs, navigate simulation parameter spaces efficiently, and quantify uncertainties.









\acknowledgments{
  This work is supported in part by the US Department of Energy SciDAC program DE-SC0021360 and DE-SC0023193, National Science Foundation Division of Information and Intelligent Systems IIS-1955764, and Los Alamos National Laboratory Contract C3435.
}

\bibliographystyle{abbrv-doi-hyperref}

\bibliography{template}

\appendix 

\end{document}